# The Reasoning Tax: Token Economics of LLM Reasoning Across Task Types and Deployment Contexts

Sachin Gopal Wani, Ajay Dholakia, and David Ellison

Lenovo, Infrastructure Solutions Group, Morrisville, NC, USA

{swani1,adholakia,dellison}@lenovo.com

**Abstract.** Accuracy-only benchmarking of reasoning-capable large language models misses a central deployment question: when do extended thinking tokens earn their cost? We introduce the Token Economy Score (TES), a marginal benchmarking metric that measures the accuracy gain of a reasoning model over a non-reasoning baseline, normalized by the generated-token multiplier. We define paired and approximated TES variants for model families with reasoning toggles and frontier models without direct non-reasoning counterparts. We then conduct an empirical benchmarking analysis across 151 model-benchmark evaluation runs on seven benchmarks spanning mathematics, code generation, science reasoning, instruction following, expert knowledge, knowledge recall, and research-level physics. The analysis examines three deployment-facing dimensions: which task structures yield positive marginal reasoning efficiency, how increasing reasoning effort changes TES within model families, and how deployment context changes economic viability. Results show that task structure predicts reasoning efficiency better than nominal difficulty: sequential inference-chain tasks such as AIME 2025 and LiveCodeBench show high TES, while knowledge-recall tasks such as MMLU-Pro show low TES despite their difficulty. We also find systematic diminishing returns at higher reasoning effort levels, including cases where additional thinking reduces accuracy. Finally, Reasoning Cost Share (RCS) shows that inference spend is often dominated by internal thinking, while Deployment Cost Multiplier (DCM) shows how on-premises deployment can change the economics of otherwise costly reasoning workloads. These findings support a benchmarking-driven model-selection rule: enable reasoning selectively by task type, effort level, and deployment context rather than treating it as a universally beneficial mode.

## 1 Introduction

Large language model inference has bifurcated into instruction-following models and reasoning-enabled models that generate extended thinking chains before answering.

Data, code, and figures available at https://github.com/Sachin-Wani/reasoning_tax

These models have raised the performance ceiling on mathematics, coding, and scientific reasoning benchmarks, but their gains come at a substantial and often unpredictable token cost: reasoning models may generate orders of magnitude more tokens than instruction-following models on the same benchmark, and a practitioner must choose between variants without knowing whether the added tokens buy enough accuracy. The decision is further complicated by deployment context, since cloud API pricing and on-premises inference can produce very different cost profiles for the same model.

This paper contributes: (1) Token Economy Score, a marginal metric for measuring whether reasoning accuracy gains justify generated-token overhead; (2) an empirical analysis of 151 model-benchmark runs across seven benchmarks, showing that sequential inference tasks yield higher reasoning efficiency than recall-heavy or saturated tasks; and (3) a deployment analysis using Reasoning Cost Share and Deployment Cost Multiplier to connect token behavior with cloud and on-premises cost.

The empirical analysis is organized around three questions: (RQ1) On which task structures do reasoning tokens provide positive marginal efficiency? (RQ2) How does increasing reasoning effort affect marginal TES within the same model family? (RQ3) How does deployment context change the economic viability of reasoning-enabled models?

## 2 Related Work

### 2.1 Reasoning Models and Chain-of-Thought Inference

Chain-of-thought prompting [9] established that eliciting intermediate reasoning steps substantially improves LLM performance on multi-step problems. Subsequent work scaled this insight into dedicated training objectives. OpenAI o1 [1] introduced reinforcement-learning-trained extended thinking, demonstrating that allowing models to generate extended internal reasoning chains before responding yields large accuracy improvements on mathematics and coding benchmarks. DeepSeek-R1 [2] brought this capability to open-weight models, achieving competitive performance with frontier proprietary systems. Qwen3 [3] introduced a built-in thinking toggle that enables direct comparison of reasoning and non-reasoning modes within a single model architecture, a property central to the TES-Δ experiments in this paper. Claude with extended thinking [4] and the Gemini 3 series [5] represent the current proprietary reasoning frontier.

### 2.2 Token Efficiency Benchmarking

The most directly related prior work is OckBench [7], which introduces Per-Token Intelligence as an absolute efficiency metric, accuracy divided by decoding token count, and identifies the Overthinking Tax, whereby smaller reasoning models generate disproportionately long reasoning chains to compensate for lower model capacity. Recent work on LLMThinkBench [22] further studies the accuracy-efficiency trade-off in basic mathematical reasoning, showing that extended reasoning can produce substantially longer outputs without monotonic accuracy improvement. These results reinforce the need for token-aware reasoning benchmarks, but focus primarily on absolute accuracy-verbosity trade-offs within basic math tasks. TES differs from this line of work in two fundamental respects. First, TES is a marginal metric: it evaluates the accuracy gain over a non-reasoning baseline rather than absolute accuracy per token. Second, TES uses generated tokens exclusively, isolating the reasoning overhead from input-token variation caused by differing evaluation harness configurations. ReEfBench [10] proposes a logical-depth framing of reasoning efficiency, modeling it as reasoning gain per unit of computational consumption. Our empirical findings complement these prior efficiency framings with a deployment-aware, task-type-stratified analysis across seven benchmarks. The growing financial burden of benchmarking frontier reasoning models has been documented in [6], which reports costs exceeding $2,700 to evaluate a single reasoning model across standard benchmark suites, motivating the need for principled efficiency metrics beyond accuracy.

### 2.3 Inference Cost and Deployment Economics

Artificial Analysis [18] provides continuous leaderboard coverage of model accuracy, generated-token counts, and API pricing. Prior work on generative-AI total cost of ownership [8] shows that inference economics can differ substantially between cloud APIs and owned hardware. This paper links these strands by evaluating reasoning efficiency and deployment cost jointly: TES measures the marginal token efficiency of reasoning, while DCM estimates how deployment context changes the dollar cost of achieving a given TES value. Prior work by the authors on benchmarking distilled language models in resource-constrained settings [19] established the importance of efficiency metrics beyond accuracy for practical deployment, motivating the extension of that efficiency lens to reasoning-capable models.

## 3 Token Economy Score

### 3.1 Motivation

Existing efficiency metrics usually measure absolute performance, such as accuracy per generated token for a single model. TES instead measures the marginal return from enabling or selecting reasoning: how much accuracy is gained relative to the additional generated tokens consumed over a non-reasoning baseline. This matches the deployment decision faced by practitioners.

TES uses generated tokens only, defined as reasoning tokens plus final output tokens. Input tokens are excluded because they are determined primarily by the benchmark prompt and evaluation harness rather than by the model's reasoning behavior.

### 3.2 Formal Definition

Let $M_r$ denote a reasoning-enabled model and $M_b$ denote a non-reasoning baseline. Let $\text{Acc}(M, T)$ denote the accuracy of model $M$ on benchmark $T$ expressed as a percentage, and let $\text{GenTok}(M, T)$ denote the mean generated token count, defined as:

$$\text{GenTok}(M,T) = \text{ReasoningTokens}(M,T) + \text{OutputTokens}(M,T)$$

The Token Economy Score is then:

$$\text{TES}(M_r, M_b, T) = \frac{\text{Acc}(M_r, T) - \text{Acc}(M_b, T)}{\text{GenTok}(M_r, T)/\text{GenTok}(M_b, T)}$$

The numerator expresses the accuracy gain in percentage points. The denominator expresses the generated token cost as a multiplier over the baseline. The TES scale is a practitioner-centered convention rather than a universal utility function. We treat one percentage point of accuracy gain as the breakeven unit against one unit of generated-token multiplier: for example, a five-point accuracy gain at a 5x generated-token multiplier yields TES = 1. Applications with higher or lower value per accuracy point can adjust this threshold, but TES = 1 provides a consistent default for comparing reasoning deployments across benchmarks.

The resulting metric has the following interpretation:

- $\text{TES} > 1$: The accuracy gain in percentage points exceeds the token cost multiplier. Reasoning is highly efficient on this task.
- $0 < \text{TES} \leq 1$: The reasoning model improves on the baseline but the accuracy gain is smaller than the token cost multiplier. Reasoning is marginal on this task.
- $\text{TES} \leq 0$: The reasoning model matches or underperforms the baseline despite higher token expenditure. Reasoning is wasteful or harmful on this task.

A negative TES is not a degenerate case. It is an empirically observed outcome, reported in Section 5, occurring when reasoning models overthink problems that instruction-following models resolve correctly without extended deliberation.

TES should be interpreted together with baseline accuracy and absolute accuracy. Near-ceiling baselines compress the numerator, while near-floor baselines can make gains appear efficient even when final accuracy remains practically low. We therefore report absolute accuracy alongside TES for frontier-difficulty regimes and leave headroom-normalized TES to future work.

### 3.3 TES Variants

Both variants share the same formula and differ only in how the baseline $M_b$ is selected.

**TES-Δ (Paired Delta).** TES-Δ pairs a reasoning and non-reasoning variant from the same model family. When the same weights expose a reasoning toggle, as in Qwen-style settings [3], this provides an architecturally controlled comparison. For closed-source families, TES-Δ should be interpreted as a vendor-paired comparison rather than proof that reasoning is the only changed variable.

**TES-A (Approximated Baseline).** $M_b$ is the highest-performing non-reasoning model on benchmark $T$ at the time of evaluation. Both the accuracy and the mean generated token count of $M_b$ are taken from this best instruct model, ensuring the denominator represents the realistic cost of the best available non-reasoning alternative. TES-A is applied to Gemini family models in this study, for which no non-reasoning variant is available.

TES-A conflates the reasoning mechanism with general capability differences between model families. This is a deliberate design choice aligned with the deployment decision context: a practitioner choosing between models does not have access to a hypothetical reasoning-disabled version of a frontier model. The relevant counterfactual is the best available instruct alternative. We note this explicitly as a limitation in Section 8.

### 3.4 Reasoning Cost Share

Reasoning Cost Share (RCS) is a descriptive quantity measuring the fraction of total inference cost consumed by the thinking chain:

$$\mathrm{RCS}(M_r, T) = \frac{\mathrm{CostReasoning}(M_r, T)}{\mathrm{CostTotal}(M_r, T)}$$

where $\mathrm{CostReasoning} = (\mathrm{ReasoningTokens}/10^6) \times P_{\mathrm{out}}$, with $P_{\mathrm{out}}$ the provider's per-million output token price, and CostTotal is the sum of input, reasoning, and output token costs at published API pricing. RCS is not a decision metric; it does not enter the TES formula. It is reported as an empirical observation characterizing how reasoning-dominated modern inference has become. RCS lies in [0,1) where value approaching 1.0 indicates that nearly all inference spend is consumed by the thinking chain rather than by the final answer, a condition with direct implications for agentic systems discussed in Section 7.

### 3.5 Deployment Cost Multiplier

The Deployment Cost Multiplier (DCM) is a direct empirical ratio comparing the total inference cost of a model workload under two deployment contexts:

$$\mathrm{DCM}(M, T) = \frac{\mathrm{CostTotal}(M, T, \text{cloud})}{\mathrm{CostTotal}(M, T, \text{on-prem})}$$

DCM is reported separately from TES because TES is token-ratio based, whereas DCM reflects dollar-cost differences across deployment contexts. In Section 6 we use DCM to compare the absolute cost of achieving a given TES value under cloud and on-premises deployment. On-premises costs in this study are derived from self-run evaluations on an 8xB300 system, with per-token costs calculated by dividing total system cost (amortized capital expenditure plus operating expenditure) by measured throughput in tokens per second, following the methodology in [8].

# 4 Experimental Setup

## 4.1 Benchmark Selection

We evaluate TES across seven benchmarks selected to span four distinct task types and a wide difficulty range. The selection criteria were: (1) availability of paired reasoning and non-reasoning model, enabling TES-Δ computation; (2) absence of full saturation, ensuring a non-trivial accuracy delta exists between reasoning and non-reasoning variants for at least a subset of models; (3) domain diversity, covering mathematics, science, code generation, instruction following, and knowledge recall; and (4) partial or absent saturation on non-reasoning baselines, since saturated tasks compress the TES numerator toward zero and extremely low-ceiling tasks require reporting absolute accuracy alongside TES. Both regimes are included in our benchmark set to allow cross-regime comparison.

**Table 1.** Benchmark summary. † IFBench includes 11 total pairs; aggregate statistics exclude one token-count anomaly described in Section 4.5.

| Benchmark | Task Structure | Questions | Difficulty | Paired models |
|---|---|---|---|---|
| IFBench [16] | Instruction following | 58 | Medium | 11† |
| MMLU-Pro [20] | Knowledge recall | 12000 | Medium–Hard | 10 |
| GPQA Diamond [12] | Knowledge recall | 198 | Hard | 12 |
| AIME 2025 [13] | Sequential Inference | 30 | Hard | 9 |
| LiveCodeBench [15] | Sequential Inference | Variable | Medium–Hard | 10 |
| HLE [14] | Frontier-domain reasoning | 2500 | Very Hard | 12 |
| CritPt [17] | Frontier-domain reasoning | 71 | Extreme | 13 |

The selected benchmarks cover four task structures relevant to reasoning deployment: instruction following, knowledge recall, sequential inference, and frontier-domain reasoning. We include both partially saturated benchmarks, where reasoning and non-reasoning models are close in accuracy, and low-ceiling benchmarks, where even frontier models remain far from human-level performance. This range allows TES to distinguish task structure from nominal difficulty.

## 4.2 Model Selection

We evaluate 27 distinct model configurations across the seven benchmarks, drawn from eight model families: GPT, Claude, DeepSeek, Qwen, Gemini, Grok, GLM, and Gemma. Model selection followed three criteria: paired coverage, reasoning-effort diversity, and scale diversity. Paired coverage enables TES-Δ comparisons within model families. Effort diversity allows analysis of medium, high, and maximum reasoning settings where available. Scale diversity includes both large mixture-of-experts models and smaller dense models to test whether TES trends hold across architecture and size.

### 4.3 Data Collection

Token counts and accuracy scores for cloud-hosted models were obtained from Artificial Analysis [18], which reports mean input tokens, reasoning tokens, and output tokens separately per model per benchmark. Generated tokens are computed as reasoning tokens plus output tokens, following the definition in Section 3.1. Pricing data (price per million input tokens and price per million output tokens) was also obtained at the time of data collection and is used for RCS and DCM calculations. Prices reflect published API rates as of late May 2026 and may differ from rates at the time of reading.

For four benchmarks (IFBench, GPQA Diamond, AIME 2025, and HLE) we conducted self-run evaluations for selected open-weight models (Qwen3.5-397B-A17B, Qwen3-235B-A22B, Qwen3-32B, Qwen3.6-27B, Gemma 4 31B, Gemma 4 26B-A4B) to obtain on-premises cost measurements and to validate Artificial Analysis figures against independent runs. Self-run accuracy scores were consistent with Artificial Analysis reported values within expected variance. For AIME 2025, supplementary accuracy data for Claude Opus 4.5 and Claude Sonnet 4.5 was obtained from MathArena [21], which independently evaluates models on mathematical competition benchmarks with per-problem cost tracking.

Each evaluation row records model identity, benchmark, reasoning flag, accuracy, generated-token components, price assumptions, total cost, and source. The full schema and all per-model values are provided in the public repository.

### 4.4 On-Premises Evaluation Setup

On-premises evaluations were conducted on a system equipped with 8xNVIDIA B300 GPUs, running all models in FP16 precision. Per-token cost was derived from a total cost of ownership model for the hardware system. The total system cost over a five-year operational period, comprising amortized capital expenditure and operational expenditure including power, cooling, and maintenance, amounts to $1.013 million, yielding an amortized running cost of $0.00633 per second ($0.38 per minute). Per-token cost for each model was then calculated as:

$$P_{on_prem} = \frac{C_{per_second}}{R_{throughput}/10^6}$$

where $C_{per_second}$ is the amortized system cost per second and $R_{throughput}$ is the measured token generation throughput in tokens per second. Throughput measurements capture total generation time including both prefill and decode phases, averaged across benchmark evaluation runs.

These costs assume sufficient workload volume to amortize the system over the measured serving throughput. For bursty or low-utilization deployments, effective per-token cost will be higher, and cloud APIs may remain preferable despite a large DCM under full-utilization assumptions.

### 4.5 Evaluation Validity and Limitations

First, token counts represent means across benchmark problems. Individual problem token counts vary, and the mean may be influenced by outliers, particularly for reasoning models on hard benchmarks where a small number of problems may trigger extremely long thinking chains. Mean and standard deviation of TES are visualized in Section 5 to reflect this variance.

Second, one IFBench pair was excluded from aggregate statistics. The non-reasoning variant of Grok 4.20 0309 generated 870,000 output tokens, compared with 48,000 output tokens for the reasoning-enabled counterpart. Because this reverses the expected token relationship and likely reflects a model failure mode: such as failing to trigger a stop token, entering an infinite repetitive generation loop, or experiencing a catastrophic formatting failure, rather than a meaningful non-reasoning baseline, the pair is retained in the repository but excluded from aggregate IFBench statistics.

Third, Gemini family models are evaluated under TES-A only due to the absence of non-reasoning Gemini variants. This means Gemini TES values reflect a comparison against the best available non-reasoning model rather than a paired architectural comparison and should be interpreted accordingly.

Finally, pricing-dependent quantities such as RCS, cloud cost, and DCM reflect the API prices available at the time of data collection; TES itself is token-ratio based and does not depend on prices.

## 5 Empirical Analysis

### 5.1 TES Across Benchmarks

The public repository reports the full pairwise TES dataset; Figure 1 summarizes the main aggregate patterns. The central finding is that task structure governs TES more strongly than raw task difficulty. We use task structure operationally: sequential inference-chain tasks require intermediate steps whose outputs constrain later steps, such as proof construction, program synthesis, debugging, or multi-constraint reasoning. Recall-heavy tasks primarily require retrieving or recognizing existing knowledge, so additional thinking cannot reliably recover missing facts. Saturated tasks compress TES because strong non-reasoning baselines leave little accuracy headroom, while near-floor tasks require reporting absolute accuracy alongside TES. Under this taxonomy, AIME 2025 and IFBench show the highest mean TES, while MMLU-Pro [11, 20] and GPQA Diamond show lower TES despite being comparably more difficult by conventional academic standards. Pair counts per benchmark range from 9 (AIME 2025) to 13 (CritPt); see repository for full per-model results.

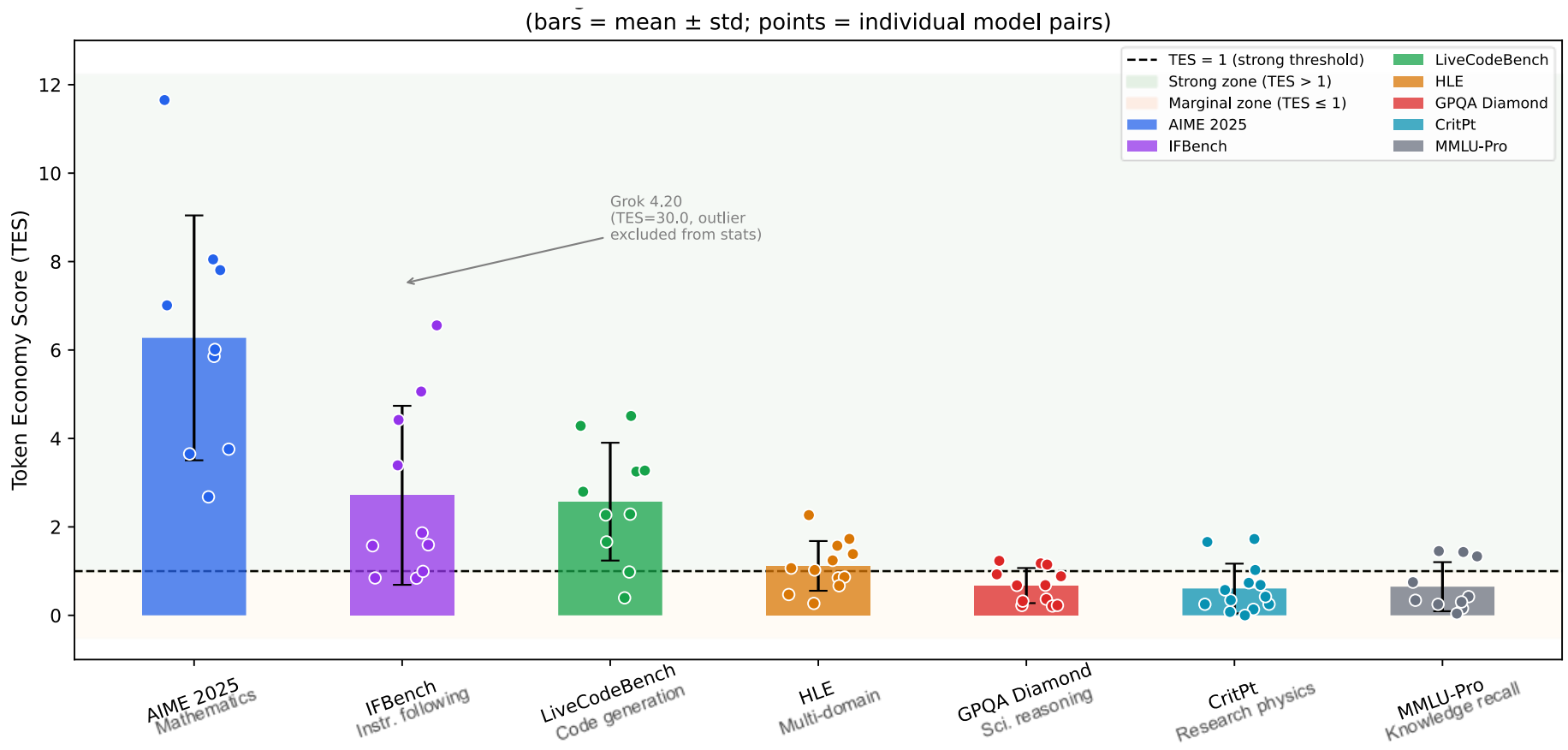

**Fig. 1.** TES distribution across benchmarks.

Figure 1 shows that TES varies more by task structure than by nominal difficulty. Sequential inference-chain tasks produce the strongest returns: AIME 2025 has the highest mean TES, while LiveCodeBench and IFBench also exceed the TES > 1 threshold on average. These tasks require intermediate steps whose outputs constrain later actions, such as proof construction, program synthesis, debugging, or multi-constraint instruction following. In contrast, MMLU-Pro and GPQA Diamond show weak TES despite high difficulty, because factual recall, saturation, or limited headroom restrict the marginal value of additional thinking. HLE and CritPt represent frontier-difficulty regimes where TES must be interpreted with absolute accuracy: HLE shows marginally positive returns, while CritPt compares reasoning models against near-floor baselines.

Across benchmarks, these results point to a single operational rule: enable reasoning selectively by task structure, not by difficulty. Sequential inference tasks benefit most; recall-heavy and saturated tasks do not. Frontier-difficulty tasks require reporting absolute accuracy alongside TES.

Gemini results are reported under TES-A and should be interpreted as approximated cross-family comparisons.

### 5.2 Diminishing Returns Within Reasoning Effort Levels

A distinct and practically important finding emerges from model families that expose multiple reasoning effort levels. Across every such family in the dataset, increasing reasoning budget beyond a moderate level yields sharply diminishing marginal TES, and in some cases, negative marginal accuracy.

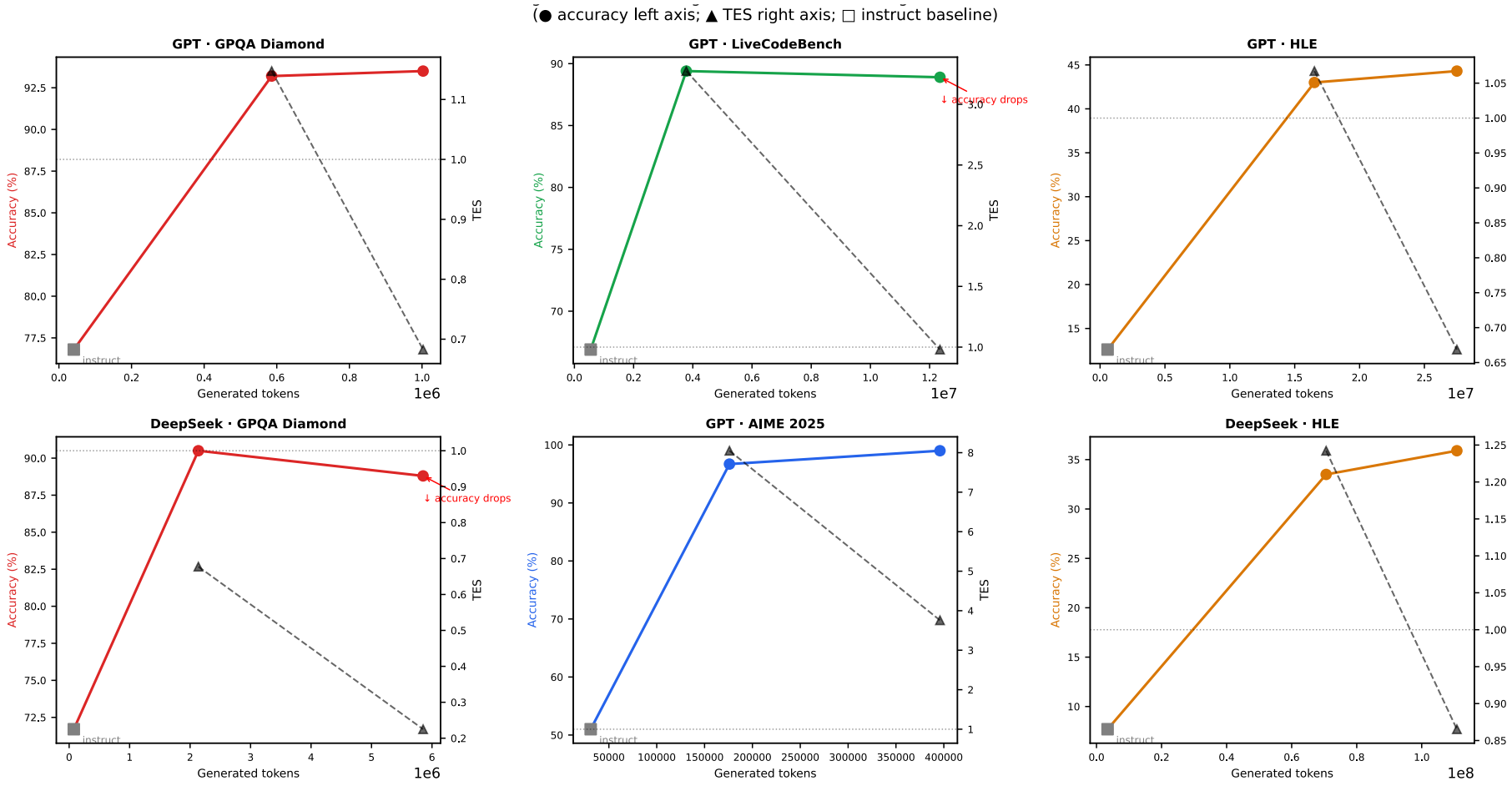

**Fig. 2.** Diminishing returns in reasoning effort.

**Table 2.** Effort-level transitions and marginal TES. TES values are computed relative to the non-reasoning baseline; marginal TES measures the efficiency of the transition to the next effort level. ↓ indicates cases where accuracy declines as reasoning budget increases.

| Model family | Benchmark | Transition | Acc change (pp) | TES (lower effort) | TES (higher effort) | Marginal TES |
|---|---|---|---|---|---|---|
| GPT-5.5 | GPQA Diamond | High → Xhigh | +0.3 | 1.147 | 0.683 | 0.175 |
| GPT-5.5 | HLE | High → Xhigh | +1.3 | 1.067 | 0.669 | 0.781 |
| GPT-5.2 | MMLU-Pro | Medium → Xhigh | +1.5 | 1.333 | 0.303 | 0.256 |
| GPT-5.2 | AIME 2025 | Medium → Xhigh | +2.3 | 8.049 | 3.758 | 1.022 |
| GPT-5.2 | LiveCodeBench | Medium → Xhigh | −0.5 ↓ | 3.274 | 0.980 | −0.153 |
| DeepSeek V4 Pro | GPQA Diamond | High → Max | −1.7 ↓ | 0.677 | 0.225 | −0.621 |
| DeepSeek V4 Pro | HLE | High → Max | +2.4 | 1.242 | 0.865 | 1.530 |

Three patterns emerge from the effort-level analysis. First, TES declines as reasoning effort increases, even when accuracy improves, because additional tokens grow faster than the accuracy gain. Second, inference-chain tasks tolerate higher effort better than recall-heavy tasks: AIME remains strongly positive even at xhigh effort, whereas MMLU-Pro collapses toward marginal TES. Third, maximum effort can reduce accuracy, as observed for DeepSeek V4 Pro on GPQA Diamond and GPT-5.2 on LiveCodeBench. These cases are consistent with an overthinking failure mode: once the model has passed a task-specific reasoning saturation point, additional generated tokens may introduce spurious alternatives, instruction drift, or verification noise rather than improving the final answer.

The practical implication is a test-and-scale workflow for reasoning effort. Start with medium or high effort on a representative validation set, measure marginal TES against the next effort level, and increase effort only when the additional accuracy gain justifies the added token cost and latency. Maximum effort should be treated as a task- and model-specific setting requiring evidence, not as the natural default for hard tasks.

### 5.3 Reasoning Cost Share

Figure 3 shows the cost composition of reasoning model inference across benchmarks, decomposed into input token cost, reasoning token cost, and output token cost.

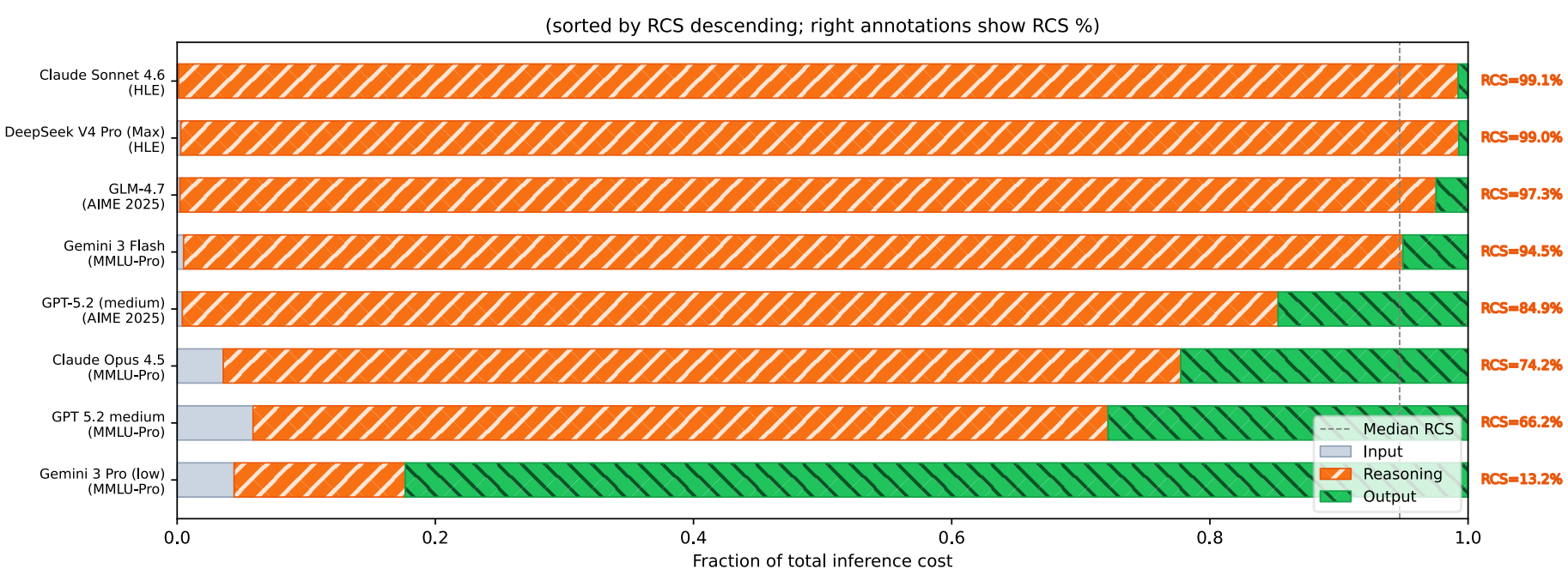

**Fig. 3.** Reasoning Cost Share across representative reasoning-enabled runs.

Reasoning tokens dominate inference cost in most reasoning-enabled runs. Across all reasoning entries, median RCS is 94.7%, and several frontier-model runs exceed 99%, meaning that nearly all generated-token spend is allocated to internal thinking rather than final answer text. Low-RCS cases occur primarily at low reasoning effort or on tasks with relatively verbose final outputs.

RCS also increases with reasoning effort. For example, GPT-5.5 on GPQA Diamond shifts from 76.7% RCS at high effort to 85.5% at xhigh effort. Combined with the diminishing-return results in Section 5.2, this shows that higher effort often reallocates more generated-token cost toward thinking while producing only small accuracy gains. Practically, high-RCS models are appropriate when hidden deliberation produces measurable accuracy gains, but inefficient when the task only requires retrieval, routing, formatting, or short final answers.

## 6 Deployment Analysis

### 6.1 On-Premises vs Cloud API

The TES values reported in Section 5 are computed using generated token ratios and are therefore independent of pricing. However, the absolute inference cost of a given TES value depends critically on where the model runs. The same reasoning model serving the same workload can cost 2x to 26x less on owned hardware than on a cloud API, depending on model architecture. This section quantifies that difference and shows how it changes the practical deployment recommendation for reasoning models.

**Table 3.** Deployment Cost Multiplier (DCM) for models evaluated on-premises on the 8x NVIDIA B300 system

| Model | Arch | Cloud $/M | On-prem $/M | Throughput | DCM |
|---|---|---|---|---|---|
| Qwen3.5-397B-A17B | MoE | $3.60 | $0.142 | 44400 | 25.6x±0.6x |
| Qwen3.6-27B | Dense | $3.60 | $0.181 | 35000 | 20.0x±1.9x |
| Qwen3-235B-A22B | MoE | $2.15 | $0.162 | 39000 | 13.6x±0.3x |
| Qwen3-32B | Dense | $0.52 | $0.196 | 32300 | 2.8x±0.3x |
| Gemma 4 26B-A4B | MoE | $0.30 | $0.067 | 94000 | 4.5x±0.1x |
| Gemma 4 31B | Dense | $0.37 | $0.192 | 33000 | 2.0x±0.1x |

DCM reflects both architecture and provider pricing. Mixture-of-Experts (MoE) models benefit from sparse activation, which increases throughput per accelerator-hour and lowers on-premises cost per token. Dense models show smaller architectural gains, so their DCM is driven more by cloud API pricing. This distinction matters because DCM for newer dense models may shrink as API prices normalize, whereas MoE advantages are more directly tied to serving efficiency.

DCM values are relatively stable across benchmarks within each model. This indicates that DCM is primarily a property of model architecture, hardware configuration, and provider pricing rather than benchmark identity. For comparable workloads on the same hardware, it can therefore be treated as an approximate deployment-cost multiplier.

Because TES is token-ratio based, DCM does not change the measured TES of a model. Instead, it changes the dollar cost of achieving that TES. Thus, a model that appears marginal under cloud pricing may become economically attractive on owned hardware if its DCM is large.

Table 3 illustrates several consequential examples. Qwen3.5-397B-A17B on GPQA Diamond achieves TES of 1.233. On cloud, the benchmark run costs $4.889; on the measured on-premises system, the same workload costs $0.192. Qwen3-235B-A22B on MMLU-Pro remains marginal in token-efficiency terms, with TES of 0.340, but on-premises deployment reduces the cost per run from $92.07 to $6.85. Thus, DCM does not change whether reasoning is token-efficient, but it can change whether a recurring reasoning workload is economically viable.

### 6.2 Model Selection Implications

Synthesizing the TES findings from Section 5 with the DCM values from Section 6.1 yields a practical model selection framework. Figure 4 plots the 77 reasoning/non-reasoning TES comparisons in TES versus cloud inference cost space, with deployment recommendations emerging from four quadrants.

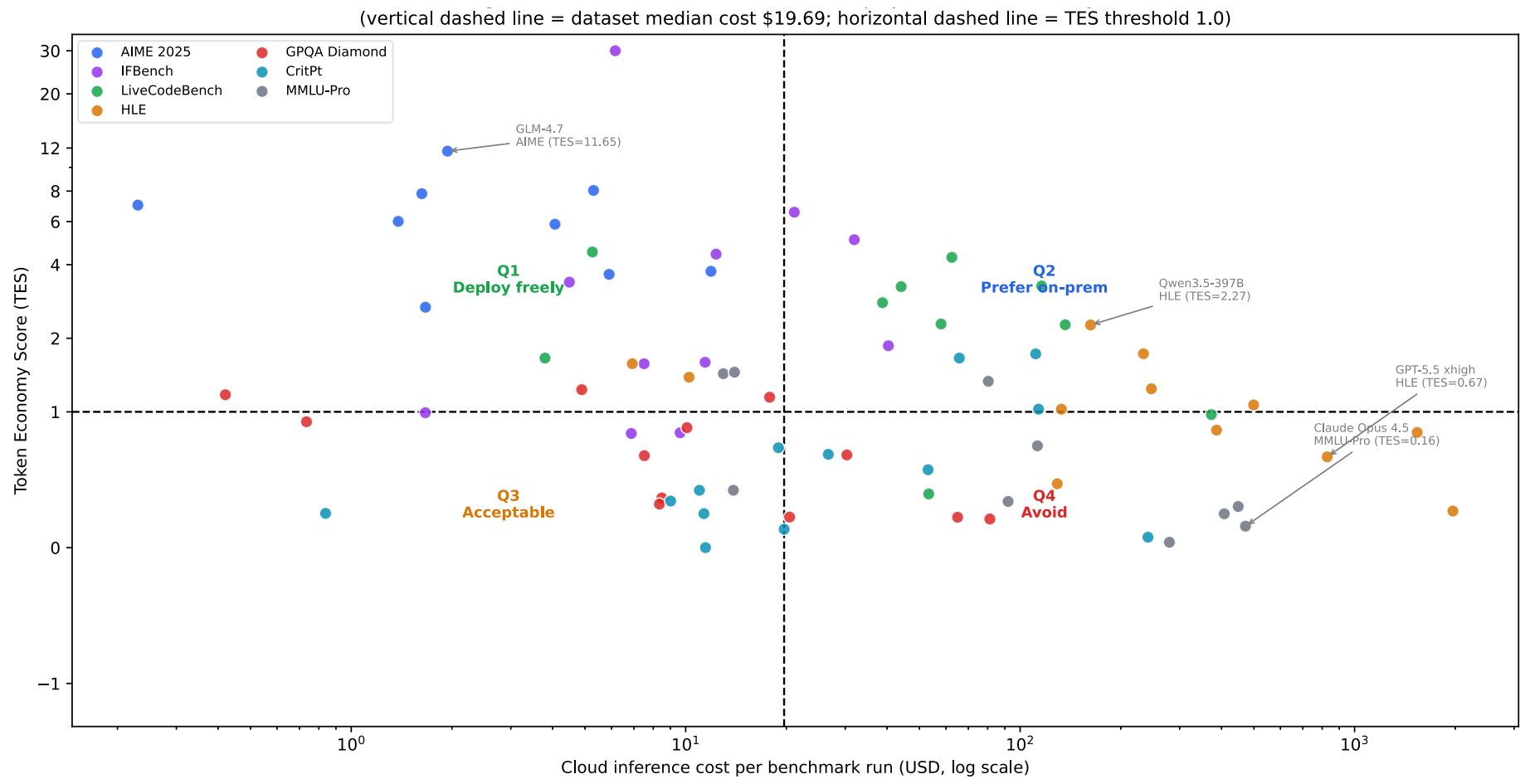

**Fig. 4.** TES vs. cloud inference cost quadrant plot.

The quadrant structure reveals four actionable deployment patterns, defined by TES > 1 as the efficiency threshold and the dataset median cloud cost of $19.69 per benchmark run as the cost threshold.

**Table 4.** Explanation of Quadrants for model selection

| Quadrant | TES/cost condition | Recommendation | Dominant examples |
|---|---|---|---|
| Q1 | High TES, low cost | Enable reasoning by default | AIME, selected Live-CodeBench |
| Q2 | High TES, high cost | Enable selectively; prefer on-prem if recurring | HLE, LiveCodeBench frontier models |
| Q3 | Low TES, low cost | Use only when the accuracy gain has application value | CritPt, some GPQA/IFBench |
| Q4 | Low TES, high cost | Disable reasoning or reduce effort | MMLU-Pro, maximum-effort HLE |

The most actionable region is Q4, where reasoning adds substantial cost without proportional accuracy gain. These cases are concentrated in knowledge-recall benchmarks and maximum-effort settings; the practical response is to disable reasoning, reduce effort, or replace the model with a strong non-reasoning baseline.

Q2 shows where deployment context matters most. High-TES models with high cloud cost may become practical when run on-premises, especially for MoE architectures with large DCM values and recurring workloads that justify hardware utilization.

Q1 contains almost exclusively AIME 2025 entries and selected LiveCodeBench runs at moderate effort which are the clearest cases where reasoning should be enabled by default regardless of deployment context. Q3 cases, concentrated in CritPt and some GPQA Diamond entries at low cost, represent situations where reasoning is marginally inefficient, but the absolute cost is low enough that the deployment decision should be driven by application-level accuracy requirements rather than TES alone.

# 7 Discussion: Implications for Agentic Systems

## 7.1 Token Compounding in Multi-Turn Systems

The TES analysis in Sections 5 and 6 treats each benchmark evaluation as a single-turn inference: one prompt, one response, and one generated-token count. Agentic systems violate this assumption because outputs from earlier steps may become part of the input context for later steps. Reasoning tokens therefore shift from a per-call cost to a potential pipeline-level cost.

This compounding effect depends on what is retained. For APIs that hide internal reasoning, high-RCS behavior increases cost and latency but does not necessarily increase downstream context length. When scratchpads, reasoning traces, or verbose intermediate outputs are retained, however, generated tokens from step j can become input tokens at later step (k). If each step's output is retained, the input context grows approximately as: $I_k = I_0 + \sum_{j=1}^{k-1} G_j$ ; where $G_j$ denotes generated tokens at step j. In this setting, high-RCS models can leave behind reasoning residue that helped produce a local answer but does not necessarily provide compact task state, reusable memory, or structured evidence for later steps. TES captures the local efficiency of reasoning, but not the downstream context pressure created when intermediate outputs are carried forward.

This distinction is especially important in orchestrator-subagent systems. If the orchestrator uses a high-RCS reasoning model, internal deliberation can compete with instructions, retrieved evidence, tool outputs, and subagent responses for context budget. If subagents return verbose reasoning traces, they can similarly crowd out the state needed for later planning. A reasoning strategy that is acceptable in a single-turn benchmark can therefore become inefficient at the system level.

## 7.2 Operational Guidelines for Agentic Deployment

The empirical results suggest three deployment guidelines.

**First, match reasoning to step type rather than pipeline type.** A pipeline that contains hard reasoning tasks does not require reasoning models at every step. Retrieval, routing, formatting, summarization, and routine tool calls resemble the low-TES tasks identified in Section 5. Reasoning should be reserved for steps involving planning, mathematical computation, code generation, verification, or other sequential inference processes.

**Second, treat the context window as a reasoning budget.** In long-horizon agents, the context window is shared by user instructions, retrieved evidence, tool outputs, intermediate state, and reasoning residue. A practical deployment check is to estimate the number of reasoning-enabled steps multiplied by the expected generated-token count per step. If this consumes a large fraction of the available context window, the system should reduce reasoning-enabled steps, lower effort, or summarize intermediate outputs before passing them forward.

**Third, control reasoning effort per step.** For intermediate reasoning steps, medium or high effort is usually a safer default than maximum effort because the results in Section 5.2 show sharply diminishing marginal TES at higher effort levels. Maximum effort should be reserved for terminal decisions, high-stakes verification, or steps where failure invalidates the full pipeline, and should be used only after measuring marginal gain on the target workload.

### 7.3 Toward Agentic TES

Existing agentic benchmarks [23, 24, 25] evaluate task completion but generally do not decompose token cost by step type, reasoning contribution, or context growth. Extending TES to agentic systems requires benchmark protocols that track both pipeline success and step-level token accounting. A future agentic TES metric should distinguish tokens that advance task state from tokens that merely increase context pressure and should relate local reasoning efficiency to end-to-end task completion. The single-turn TES introduced in this paper provides the per-call foundation for this analysis, but agentic evaluation requires an additional accounting layer across steps.

## 8 Conclusion

This paper introduced Token Economy Score, a marginal efficiency metric for evaluating whether the accuracy gains of reasoning-capable LLMs justify their generated-token overhead. By comparing reasoning models against non-reasoning baselines, TES frames reasoning as a deployment decision rather than an accuracy-only leaderboard outcome.

Returning to the three research questions: (RQ1) sequential inference-chain tasks (AIME 2025, LiveCodeBench, IFBench) consistently yield positive TES while recall-heavy tasks (MMLU-Pro) do not. (RQ2) increasing reasoning effort produces sharply diminishing marginal TES across every evaluated family, with confirmed accuracy decline at maximum effort for two model families. (RQ3) on-premises MoE deployment changes economically marginal workloads into viable ones through DCM values of 13–26x. The central empirical result is that reasoning efficiency is task-structured: sequential inference tasks benefit most, while recall-heavy and saturated tasks show weak returns. Increasing reasoning effort also exhibits sharp diminishing returns, showing that effort level should be selected explicitly rather than treated as a monotonic quality knob.

Deployment context further changes the practical interpretation of these results. RCS shows that reasoning tokens dominate inference cost in many frontier-model runs, while DCM shows that on-premises deployment can make some otherwise expensive reasoning workloads economically viable.

The main limitations are the use of benchmark-level mean token counts, approximated baselines for models without non-reasoning counterparts, pricing-epoch dependence for cost quantities, and the single-turn nature of TES. Future work should add per-problem uncertainty estimates, headroom-normalized TES for near-floor benchmarks, and agentic extensions that account for context growth across multi-step pipelines.